# Dense Volume-to-Volume Vascular Boundary Detection

Jameson Merkow[1], David Kriegman[1], Alison Marsden[2], and Zhuowen Tu[1]

[1] University of California, San Diego.
[2] Stanford University

**Abstract.** In this work, we present a novel 3D-Convolutional Neural Network (CNN) architecture called I2I-3D that predicts boundary location in volumetric data. Our fine-to-fine, deeply supervised framework addresses three critical issues to 3D boundary detection: (1) efficient, holistic, end-to-end volumetric label training and prediction (2) precise voxel-level prediction to capture fine scale structures prevalent in medical data and (3) directed multi-scale, multi-level feature learning. We evaluate our approach on a dataset consisting of 93 medical image volumes with a wide variety of anatomical regions and vascular structures. In the process, we also introduce HED-3D, a 3D extension of the state-of-the-art 2D edge detector (HED). We show that our deep learning approach out-performs, the current state-of-the-art in 3D vascular boundary detection (structured forests 3D), by a large margin, as well as HED applied to slices, and HED-3D while successfully localizing fine structures. With our approach, boundary detection takes about one minute on a typical 512x512x512 volume.

## 1 Introduction

The past decade has witnessed major progress in computer vision, graphics, and machine learning, due in large part to the success of technologies built around the concept of "image patches". Many patch-centric approaches fall into the category of "sliding-window" methods [11, 3, 9] that performs prediction by considering dense, overlapping windows. Patch-centric approaches limit us in terms of computational complexity and long-range modeling capabilities.

Fully convolutional neural networks (FCN) [7] achieved simultaneous performance and full image labeling. Holistically-Nested Edge Detector (HED) [13], applied this approach to image-to-image object boundary detection. HED learns multi-scale features guided by deep supervision [6] on side outputs that resolve ambiguity in boundary detection with multi-scale prediction. HED significantly improved the state-of-the-art in edge detection, and did so at a fraction of the computational cost of other CNN approaches. Another member of the FCN family, UNet [10], adapted this architecture to efficiently produce neuronal segmentations.

Volume-to-volume learning has yet to garner the same attention as image-to-image labeling. One approach applies 2D prediction schemes on images generated by traversing the volume on an anatomical plane then recombining predictions into a volume. However, volumetric features exist across three spatial dimensions, therefore it is crucial to process this data where those features exist.

The current state-of-the-art in vessel wall detection uses a 3D patch-to-patch approach along with medical imaging driven features and *a-priori* information, and a structural forest classifier [9]. In that work, the authors mitigate the computational cost of the patch-centric classifiers by using an intelligent sampling scheme and limiting prediction to certain types of vascular structures. This method side-steps the inefficiency of patch-centric classifiers but, overall, their approach still limits accurate prediction to a narrow variety of structures and anatomical regions. CNN approaches have been applied to volumetric labeling [14], but the high computational cost of these frameworks limit the ability to make accurate end-to-end volumetric predictions.

A secondary challenge lies in detecting small structures prevalent in medical volume data. In contrast to photographs, where objects and boundaries of interest are well-spaced and commonly well-sized, anatomical structures are often small, and resolution may be limited by acquisition. In fact, small anomalies are often of greater importance than the larger structures These factors manifest a unique challenge for dense labeling of medical volumes.

In this work, we propose a novel 3D-CNN architecture, I2I-3D, that perform dense volume-to-volume labeling. Specifically, we tackle three key issues in dense medical volume label prediction, (1) efficient volumetric labeling of medial data using 3D, volume-to-volume CNN architectures, (2) precise fine-to-fine and volume-to-volume labeling, (3) nested multi-scale learning. We extend the typical fine-to-coarse architecture by adding an efficient means to process high resolution features late in the network, enabling precise voxel-level prediction that benefit from coarse level guidance and nested multi-scale representations. We evaluate our approach against the state-of-the-art in vessel wall detection.

## 2 Dense Volume-to-Volume Prediction

### 2.1 Pixel Level Prediction In 2D Images

Fully convolutional neural networks [7] were among the first methods to adapt the fine-to-coarse structure to dense pixel-level prediction. The FCN architecture added element-wise summations to VGGNet [2] that link coarse resolution predictions to layers with finer strides. However, it has been shown that directly pulling features from bottom layers to the fine level is sub-optimal as the fine-level features have no coarse-level guidance [4]. HED [13] produced top accuracy on the BSDS500 dataset [8] with an alternative adaptation of VGGNet which fused several boundary responses at different resolutions with weighted aggregation. However, HED's fine-to-coarse framework leaves fundamental limitations to precise prediction and a close look at the edge responses produced by HED reveals many thick orphan edges. HED only achieves top accuracy after boundary refinement via non-maximum suppression (NMS) and morphological thinning. This approach is sufficient for 2D contour detection tasks, however increased degrees-of-freedom in 3D data makes NMS unreliable for volumetric data. Furthermore, NMS fails when the prediction resolution is lower than the object separation resolution.

In these architectures, the most powerful outputs, in terms of predictive power, lack the capability to produce fine resolution predictions. Not only is this problematic for making high resolution predictions, but coarse representations can inform finer resolution predictions and finer resolution distinctions often require complex predictive power. UNet [10] addressed some of these issues by adapting FCN's architecture for neuronal segmentation. Their approach replaces summation units with concatenation and multiple convolutional layers, and applies weighted loss function which penalizes poor localization of adjacent structures. Though, UNet improves localization, its large number of dense layers make it too inefficient for 3D tasks; furthermore, it does not directly learn nested multi-level interactions.

### 2.2 Precise Multi-Scale Voxel Level Prediction

Our framework addresses these critical issues to volume-to-volume labeling and applies them to vascular boundary detection in medical volumetric images. Our proposed network, I2I-3D consists of two paths, a fine-to-coarse path and a multi-scale coarse-to-fine path. The fine-to-coarse network structure follows popular network architecture and generates features with increasing feature abstraction and greater spatial extent. By adding side outputs and multi-scale fusion to this path, we obtain an efficient 3D-CNN, we dub HED-3D. However, HED-3D fails to localize small vascular structures, and we add a secondary path to increase the framework's prediction resolution. The secondary path combines fine-to-coarse features and learns complex multi-scale interactions in a coarse-to-fine fashion.

Each stage of the coarse-to-fine path incorporates abstract representations with higher resolution features and generates a new higher resolution response with multi-scale influences and coarse level guidance. Special 'mixing' layers followed by convolution layers combine these responses and minimize a multi-scale objective function. Deep supervision [6] plays an important role in our approach; it directly enforces multi-scale integration at each stage. When cascaded, this process results in features with very large projective fields at high resolution. Later layers benefit from abstract features, coarse level guidance and multi-scale integration resulting in a top most layer with the best predictive power and highest resolution. Figure 1 depicts the layer-wise connected 3D convolutional neural network architecture of I2I-3D.

### 2.3 Formulation

We denote our input training set of N volumes by $S = \{(X_n, Y_n), n = 1, \ldots, N\}$, where sample $X_n = \{x_j^{(n)}, j = 1, \ldots, |X_n|\}$ denotes the raw input volume and $Y_n = \{y_j^{(n)}, j = 1, \ldots, |X_n|\}, y_j^{(n)} \in \{1, .., K\}$ denotes the corresponding ground truth label map. For our task, $K = 2$, here we define the generic loss formulation. We drop $n$ for simplicity, as we consider volumes independently. Our goal is to learn network parameters, $\mathbf{W}$, that enable boundary detection at multiple resolutions. Our approach produces $M$ multi-scale outputs with $\frac{1}{2^{M-1}}$ resolution of input. Each output has an associated classifier whose weights are denoted $\mathbf{w} = (\mathbf{w}^{(1)}, \ldots, \mathbf{w}^{(M)})$. Loss for each of these outputs is defined as:

$$\mathcal{L}_{\text{out}}(\mathbf{W}, \mathbf{w}) = \sum_{m=1}^{M} \ell_{\text{out}}^{(m)}(\mathbf{W}, \mathbf{w}^{(m)}), \tag{1}$$

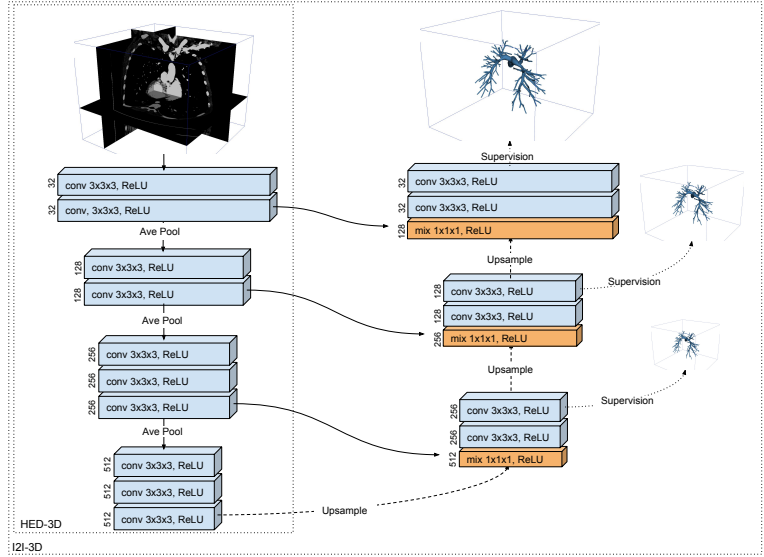

**Fig. 1:** Depiction of the proposed network architecture, I2I-3D, which couples fine-to-coarse and coarse-to-fine convolutional structures and multi-scale loss to produce dense voxel-level labels at the resolution of the input. The number of channels is denoted on the left of each convolution layer, arrows denote network connections and operations.

where $\ell_{\text{out}}$ denotes the volume-level loss function for outputs. Loss is computed over all voxels in a training volume $X$ and label map $Y$. Specifically, we define the following cross-entropy loss function used in Eqn. (1):

$$\ell_{\text{out}}^{(m)}(\mathbf{W}, \mathbf{w}^{(m)}) = -\sum_{k}\sum_{j\in Y_k} \log \Pr(y_j = k|X; \mathbf{W}, \mathbf{w}^{(m)}) \qquad (2)$$

where $Y_k$ denotes the voxel truth label sets for the $k^{th}$ class. $\Pr(y_j = k|X; \mathbf{W}, \mathbf{w}^{(m)}) = \sigma(a_j^{(m)}) \in [0, 1]$ is computed using sigmoid function $\sigma(.)$ on the activation value at voxel $j$. We obtain label map predictions $\hat{Y}_{\text{out}}^{(m)} = \sigma(\hat{A}_{\text{out}}^{(m)})$, where $\hat{A}_{\text{out}}^{(m)} \equiv \{a_j^{(m)}, \ j = 1, \ldots, |Y|\}$ are activations of the output of layer $m$. Putting everything together, we minimize the following objective function via standard stochastic gradient descent:

$$(\mathbf{W}, \mathbf{w}) = \operatorname{argmin}(\mathcal{L}_{\text{out}}(\mathbf{W}, \mathbf{w})) \qquad (3)$$

During testing, given image $X$, we obtain label map predictions from the output layers: $\hat{Y}_{\text{top}} = \text{I2I}(X, (\mathbf{W}, \mathbf{w}))$, where $\text{I2I}(\cdot)$ denotes the label maps produced by our network.

## 3 Network Architecture and Training

With 16 convolutional layers each single stride, and multiple stages, VGGnet [2] possesses great depth and density generating outstanding accuracy. We mimic a 3D version VGGNet's architecture with domain specific modifications for our HED-3D framework and the fine-to-coarse path of I2I-3D. We truncate at the fourth pooling layer and all fully connected layers are removed, resulting in a network with 10 convolutional layers and four resolutions. In addition, we decrease the filter count in the first two convolution layers to 32 and replace max pooling with average pooling. As in [13], we place deep supervision at side-outputs at each convolution layer just prior to pooling, and these side-outputs are fused via weighted aggregation.

I2I-3D adds a secondary structure to HED-3D that processes multi-scale information to produce precise labels that benefit from coarse-level guidance. The second structure follows an inverted pattern of the fine-to-coarse path that starts at the lowest resolution and upsamples in place of pooling. Each stage of the coarse-to-fine path contains a mixing layer, followed by two convolutional layers. Mixing layers take two inputs; one from the corresponding resolution in fine-to-coarse path and a second from the output of the previous (coarser) stage in the coarse-to-fine path. Mixing layers concatenate these inputs and use a specialized 1x1x1 convolution to directly mix input features. Mixing layers are similar to reduction layers in GoogLeNet[12] but differ in usage and initialization. These layers directly hybridize the lower resolution inputs with the fine-to-coarse stream, in addition to maintaining network efficiency. Two convolutional layers follow the mixing layer, and add spatial mixing of the two streams. Each stage contains a deeply supervised output at the final convolution layer. A key difference between side outputs in I2I-3D and HED-3D is that these outputs are only used to promote multi-scale integration at each stage, and are eventually phased out leaving single output at the top most layer.

We begin by describing the training procedure for HED-3D. We, first, load pre-trained weights (details in Sec 4) and place deep supervision at each of the four multi-scale outputs and at the fusion output. We literately train with a base learning rate of $1e^{-7}$ which is decimated every $30k$ iterations. Weight updates occur after every iteration till convergence (after about iterations).

I2I-3D builds upon this procedure by attaching the coarse-to-fine path and placing deep supervision at each of the new multi-scale outputs. Mixing layers are initialized to output only their fine-to-coarse feature inputs. The coarse-to-fine convolution layers are initialized to produce an identity mapping. Initially, we decrease the learning rates in the fine-to-coarse path by a factor of 100 and train until the multi-scale loss plateaus. These initializations and learning rates force the network to learn nested multi-resolution features and multi-scale integration at each stage to minimize multi-scale loss. Finally, we return the learning rate multipliers in the fine-to-coarse path to their original setting, remove all supervision except the top-most, highest resolution output, and train until convergence. This allows new information to be encoded after the coarse-to-fine outputs have assimilated multi-scale information.

## 4   Experimentation and Results

In [9], the authors use direct voxel overlap for evaluation, however this metric fails to account for any localization error in boundary prediction and over-penalizes usable boundaries that do not perfectly overlap with ground truth boundaries. Instead, we evaluate with metrics that apply the same principles as the BSDS benchmark [1] which are standard protocols for evaluating boundary contours in natural images. This metric evaluates performance by finding match correspondences between ground truth and predicted contour boundaries. Matched voxels contribute to true positive counts, and unmatched voxels contribute to fall-out and miss rates. We extended these metrics[1] to three spatial dimensions and we report three performance measures: fixed threshold F measure (ODS), best per-image threshold F measure (OIS), and average precision (AP) as well as precision-recall curves for each classifier.

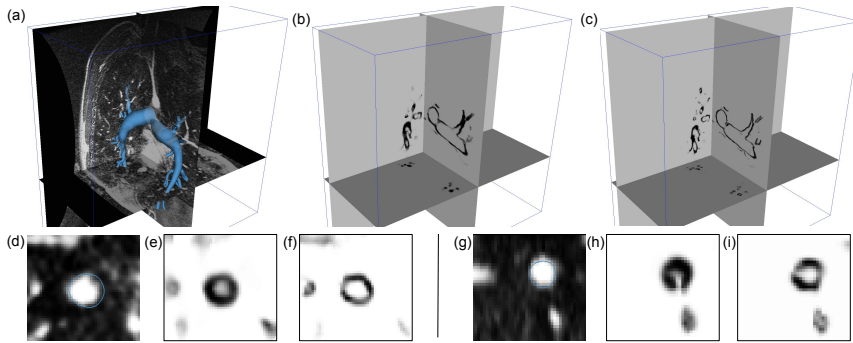

**Fig. 2:** Results of our HED-3D and I2I-3D vessel boundary classifiers. (a) Input volume and ground truth (in blue). (b) HED-3D result. (c) I2I-3D result. (d),(g) vessel cross section and ground truth (in blue). (e),(h) HED-3D cross section result. (f),(i) I2I-3D cross section result.

We evaluate our method on a dataset consisting of 93 medical image volumes which was obtained from the vascular model repository[2]. Our dataset includes both magnetic resonance (MR) and computed tomography (CT) volumes each with 3D models built for computational blood flow simulation. We expanded the dataset used in [9] by adding new volumes to include a wide variety of regions: abdominal, thoracic, cerebral, vertebral, and lower extremity regions. Each volume was captured from individual patients for a wide variety of clinically indicated purposes, including aneurysm, stenosis, peripheral artery disease, congenital heart disease, and dissection, as well as patients with normal physiology. All arterial vessel types except coronary were expertly annotated however only one structure is annotated per volume. Since each volume contains only selected ground truth annotation, we only consider areas within 20 voxels of a vessel during evaluation. This procedure does not remove any areas inside the vessel where correct predictions are critical. We split this dataset into training, validation and test sets, which contain 67, 7 and 19, volumes respectively.

Prior to network training a number of pre-processing steps are performed. We begin by whitening each individual volume in our dataset and cropping the volumes into overlapping segments. We found that 96x96x48 segments allow fast processing while maintaining a large spatial extent. A single segment takes about one second to process on one NVidia K40 GPU and a typical volume (512x512x512) takes less than a minute on the same hardware. We augmented our data by cropping overlapping segments; each segment overlaps its neighbors by 12x12x8 voxels during training. Since ground truth vessel wall are spatially sparse and labeling is inconsistent, only volumes that contain over 0.25% labeled vessel voxels (approx. 1000 of 442,368 voxels) were considered during training.

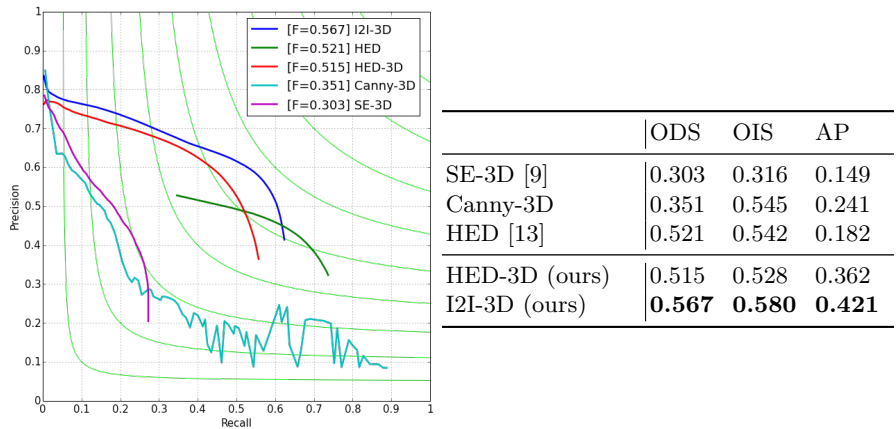

| | ODS | OIS | AP |
|---|---|---|---|
| SE-3D [9] | 0.303 | 0.316 | 0.149 |
| Canny-3D | 0.351 | 0.545 | 0.241 |
| HED [13] | 0.521 | 0.542 | 0.182 |
| HED-3D (ours) | 0.515 | 0.528 | 0.362 |
| I2I-3D (ours) | **0.567** | **0.580** | **0.421** |

**Fig. 3:** (left) Precision recall curves comparing our approach with state-of-the-art, our baseline methods. (right) Table of performance metrics of our approach and baselines.

We implemented our network using the popular *Caffe* library [5] and methods were extended for 3D where necessary[3]. The fine-to-coarse weights for all 3D-CNN networks were generated by first pre-training a randomly initialized network and on entire vessel label prediction for a fixed number of iterations ($50k$) with a high learning rate. Full vessel labels naturally accompany vessel wall labels but provide less overall loss, in turn preventing unstable gradients during back-propagation.

We compare I2I-3D to the current state of the art [9], and a 2D-CNN baselines as well as our HED-3D architecture. HED (in 2D) was trained without modification on 2D slices of each volume generated by navigating along an anatomical axes. We also compare against the widely used 3D-Canny edge detector.

Fig. 2 and Fig. 3 show a comparison of results. Fig 3 indicates that our method out-performs the current state-of-the-art by a wide margin. We also notice the benefit of 3D-CNN over 2D-CNN architectures when comparing results of HED and HED-3D. We observe that I2I-3D shows a large improvement over HED-3D, indicating that fine-to-fine multi-scale learning significantly improves predictive capabilities. In Fig. 2, we see the results of I2I-3D characterized by stronger and more localized responses when compared to HED-3D, showing the benefit of our fine-to-fine, multi-scale learning approach. By using a fine-to-coarse architecture, both HED and HED-3D produce lower resolution results which are incapable of modeling tiny structures. This results in poor localization and a weaker edge response. By learning the multi-scale interaction, I2I-3D is able to precisely localize vessel boundaries.

---

[3] Our code will be released upon publication

## 5    Conclusion

In this work, we present a 3D-CNN approach that address major issues in efficient volume-to-volume labeling. Our HED-3D framework demonstrates that processing volumetric data natively, in 3D, leads to gains in performance over a 2D counterpart. In addition, our framework, I2I-3D, is a computationally efficient means for learning multi-scale hierarchal features. We use these features to generate precise voxel-level predictions at input resolution. We demonstrate, through experimentation, that our approach is capable of fine localization by achieving state of the art performance in 3D vessel boundary detection, even without explicit *a-priori* information. We provide our 3D implementation to ensure that our approach can be applied to variety of medical applications and domains[4].

---

[4] Our code and models will be made publicly available upon publication.